\pdfoutput=1

\documentclass[11pt]{article}

\usepackage[]{EMNLP2023}
\usepackage{graphicx} 
\usepackage{subcaption}
\usepackage{amsfonts}
\usepackage{booktabs}
\usepackage{amsmath}
\usepackage{tcolorbox}
\usepackage{multirow}
\usepackage{diagbox}

\usepackage{tikz}
\usetikzlibrary{automata,arrows}
\newcommand{\framework}{grounding acts}

\newcommand\blfootnote[1]{%
  \begingroup
  \renewcommand\thefootnote{}\footnote{#1}%
  \addtocounter{footnote}{-1}%
  \endgroup
}

\usepackage{times}
\usepackage{latexsym}

\usepackage[T1]{fontenc}

\usepackage[utf8]{inputenc}

\usepackage{microtype}

\usepackage{inconsolata}

\definecolor{c1}{cmyk}{0,0.6175,0.8848,0.1490} 
\definecolor{c2}{cmyk}{0.1127,0.6690,0,0.4431} 
\definecolor{c3}{cmyk}{0.3081,0,0.7209,0.3255} 
\definecolor{c4}{cmyk}{0.6765,0.2017,0,0.0667} 
\definecolor{c5}{cmyk}{0,0.8765,0.7099,0.3647} 
\definecolor{forestgreen}{HTML}{397727}

\newtcbox{\hlprimarytab}{on line, rounded corners, box align=base, colback=c3!10,colframe=white,size=fbox,arc=3pt, before upper=\strut, top=-2pt, bottom=-4pt, left=-2pt, right=-2pt, boxrule=0pt}
\newtcbox{\hlsecondarytab}{on line, box align=base, colback=red!10,colframe=white,size=fbox,arc=3pt, before upper=\strut, top=-2pt, bottom=-4pt, left=-2pt, right=-2pt, boxrule=0pt}

\newtcbox{\hlorangetab}{on line, box align=base, colback=orange!10,colframe=white,size=fbox,arc=3pt, before upper=\strut, top=-2pt, bottom=-4pt, left=-2pt, right=-2pt, boxrule=0pt}

\newtcbox{\hlgraytab}{on line, rounded corners, box align=base,colframe=white,size=fbox,arc=3pt, before upper=\strut, top=-2pt, bottom=-4pt, left=-2pt, right=-2pt, boxrule=0pt}

\title{Grounding Gaps in Language Model Generations}

\newcommand\coauth{$^\star$}

\author{Omar Shaikh\coauth  \hspace{0.8em}
        Kristina Gligorić\coauth \hspace{0.8em}
        Ashna Khetan  \hspace{0.8em}
        Matthias Gerstgrasser  \hspace{0.8em}\\
        \textbf{Diyi Yang} \hspace{0.8em}
        \textbf{Dan Jurafsky} \\
         Stanford University\\
         \texttt{\small \{oshaikh, gligoric, ashnak, mgerst, diyiy, jurafsky\}@stanford.edu}}

\begin{document}
\maketitle
\begin{abstract}
Effective conversation requires common ground: a shared understanding between the participants. Common ground, however, does not emerge spontaneously in conversation. Speakers and listeners work together to both identify and construct a shared basis while avoiding misunderstanding. To accomplish grounding, humans rely on a range of dialogue acts, like clarification \textit{(What do you mean?)} and acknowledgment \textit{(I understand.)}. However, it is unclear whether large language models (LLMs) generate text that reflects human grounding. To this end, we curate a set of \textit{grounding acts} and propose corresponding metrics that quantify attempted grounding. We study whether LLM generations contain grounding acts, simulating turn-taking from several dialogue datasets and comparing results to humans. We find that---compared to humans---LLMs generate language with less conversational grounding,
instead generating text that appears to simply presume common ground.
To understand the roots of the identified \textit{grounding gap}, we examine the role of instruction tuning and preference optimization, finding that training on contemporary preference data leads to a reduction in generated grounding acts. Altogether, we highlight the need for more research investigating conversational grounding in human-AI interaction. 
\end{abstract}

\blfootnote{\coauth Equal contribution.}

\section{Introduction}
In dialogue, \textbf{common ground} refers to the mutual knowledge, beliefs, and assumptions shared by participants in a conversation. This shared understanding is essential for effective communication, as it underpins the ability of individuals to interpret, predict, and respond to each other's statements and actions accurately~\cite{clark1996using}. Through each conversational turn, individuals collaborate to build common ground, preventing potential misunderstandings~\cite{clark1989contributing}. Humans therefore rely on dialogue acts like \textit{clarifying} meaning or posing information-seeking \textit{followup} questions. When individuals fail to ground, they often proactively repair misunderstandings. 

\begin{figure}
    \centering
    \includegraphics[width = \linewidth]{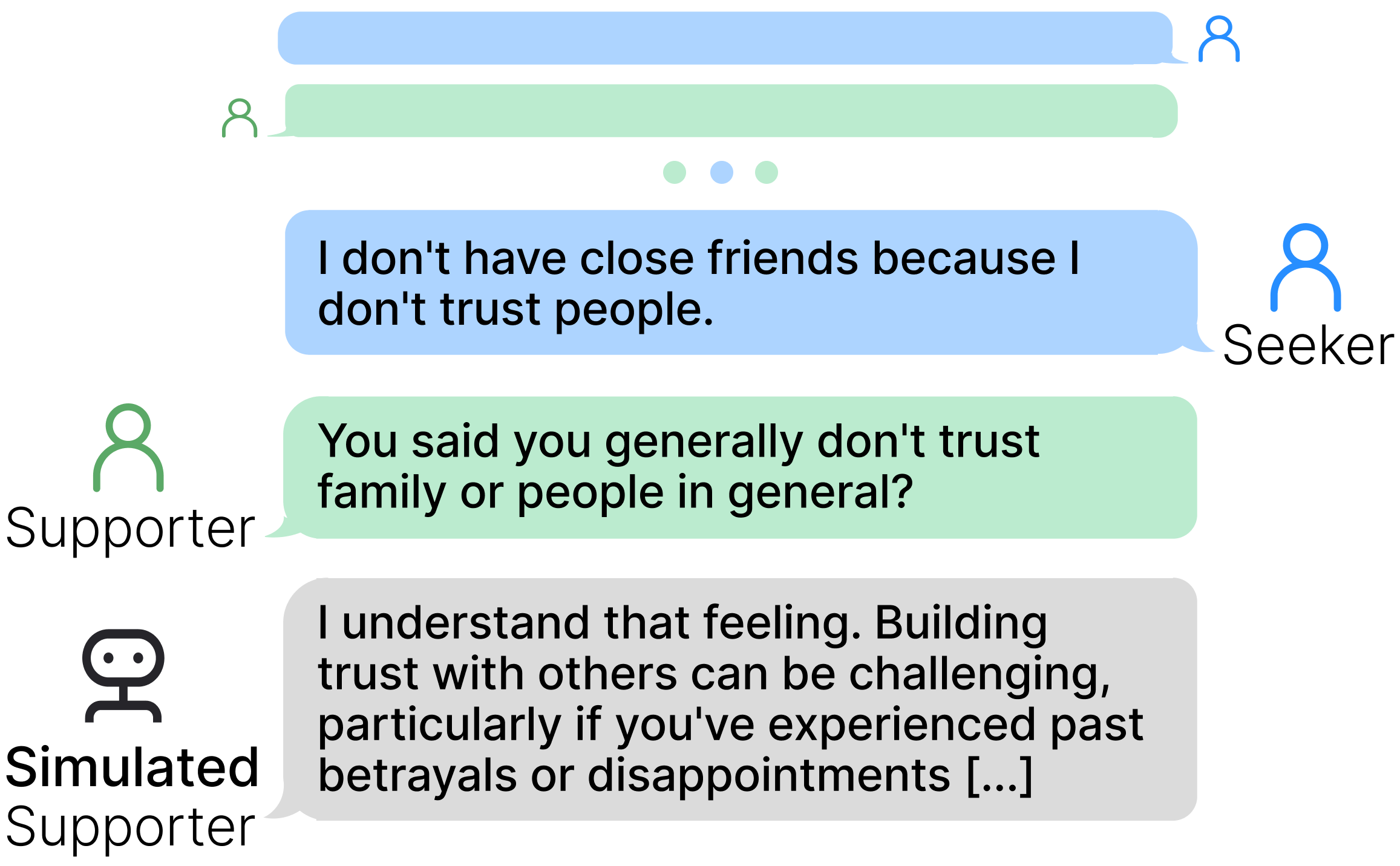}
    \caption{Mental health supporters carefully employ clarification questions (one of the three \framework{}) with a seeker, taking multiple turns to ground. In contrast, simulated supporters---with the same conversational context---generate presumptive answers. 
    }
    \label{fig:example_interact}
\end{figure}

Failing to construct common ground in human-human conversation can be at best misleading and at worst harmful. Consider mental-health support: if there's any indication of risk to a client (e.g., suicidal ideation or intentions of self-harm), a healthcare professional will ask clarifying questions to assess risk. Failure to do so has harmful consequences (unnecessary hospitalization) which can traumatize a client and place an unjustified financial burden~\cite{strumwasser1991appropriateness}. In domains like education, ineffective grounding might result in misunderstanding a student, resulting in irrelevant feedback~\cite{graesser1995collaborative}.

From an NLP perspective, establishing and leveraging common ground is a complex challenge: dialogue agents must recognize and utilize implicit aspects of human communication. Recent chat-based large language models (LLMs), however, are designed explicitly for following instructions. While humans carefully construct common ground, LLMs are trained to directly act on commands specified by end-users~\cite{ouyang2022training}. We hypothesize that the instruction-following paradigm---a combination of supervised fine-tuning (SFT) and preference optimization (PO through RLHF, for example)---might result in discrepancies between how humans actually ground in dialogue, and how LLMs generate text similar to human grounding. We call this discrepancy the \textbf{grounding gap}.

Despite the grounding gap, LLMs interact regularly with humans across various applications. For a subset of interactions, however, LLMs \textit{should} generate grounding language before completing a user's task, instead of executing literal instructions or disregarding a user's underlying goals. This is particularly crucial in LLM-powered training systems, where LLMs simulate practice scenarios and allow individuals to rehearse and refine domain-specific skills~\cite{shaikh2023rehearsal}. LLM-based training already facilitates interaction in domains like \textbf{education} \cite{kasneci2023chatgpt,demszky2021can,wang2023chatgpt}, \textbf{conflict resolution}~\cite{shaikh2023rehearsal, argyle2023ai}, and \textbf{emotional support} \cite{carlbring2023new, hsu2023helping}. In these settings, effective dialog agents must coordinate to build common ground when interacting with people.

Given the importance of generating language for conversational grounding, we ask: \emph{Do current LLMs generate dialogue acts that reflect grounding patterns between humans?} \emph{If not, what aspect of LLM training exacerbates the grounding gap?}

We address these questions by measuring LLM generations through linguistically validated \textbf{grounding acts}. For example, acts that \textit{clarify} or \textit{acknowledge} a prior utterance offer a strong signal for measuring shared understanding~\cite{clark1989contributing}. Building on prior work in dialogue and conversational analysis, we curate a collection of dialogue acts used to construct common ground (\S\ref{grounding_acts}). Then, we select datasets \& domains to study human-LM grounding. We focus on settings where human-human grounding is critical, and where LLMs have been applied: namely, emotional support, persuasion, and teaching (\S \ref{simulation_datasets}). 

After curating a set of grounding acts, we build prompted few-shot classifiers to detect them (\S \ref{simulation_compare}). We then use LLMs to simulate turn-taking in our human-human dialogue datasets and compare agreement between human and \texttt{GPT}-generated grounding strategies (\S \ref{sec:simulation}). 

Because we use the exact same conversational context as in human conversations, we can quantify the \textbf{grounding gap}: off-the-shelf LLM generations are, on average, \textbf{77.5\%} less likely to contain \framework{}  than humans (\S\ref{gaps_in_grounding}). Even in situations where LLM generations do contain a grounding act, they differ from human generations---we observe poor human-LM agreement across a range of models. 

To isolate potential causes of the grounding gap, we explore a range of possible interventions, from ablating training iterations on instruction following data (SFT and PO) to designing a simple prompting mitigation  (\S\ref{sec:ablate-instr}). We find that SFT does not improve conversational grounding, and PO erodes it. Across our experiments, we generally observe significant disagreement between \framework{} in human utterances and LLM generations. 

In summary, we contribute (1) a set of linguistically informed grounding acts, tailored towards understanding grounding in contemporary LLMs. Using grounding acts, we (2) conduct a controlled LLM simulation of conversational grounding and a characterization of \textit{grounding agreement}: a comparison between \framework{} in human utterances and LLM generations. Finally, we (3) investigate reasons for the grounding gap, exploring different models, ablating SFT and PO, and testing prompting interventions.\footnote{Code and prompts for evaluation can be found at \url{https://github.com/oshaikh13/grounding-gaps}}

\section{Grounding Acts}
\label{grounding_acts}
Grounding acts are a set of dialogue acts used to characterize conversational grounding. While grounding failures are qualitatively well-observed in current pretrained models \cite{benotti2021grounding, fried2022pragmatics, chandu2021grounding}, measuring \textit{where or how frequently} these failures occur in dialogue remains unexplored. Here, we outline concrete acts to measure these failures, discuss our curation process for selecting grounding acts, and describe each act in detail. 

\begin{table*}[]
    \centering
    \resizebox{\textwidth}{!}{
    \begin{tabular}{l|l}
        \toprule
        Grounding Act & Example \\
        \midrule
        Clarification & \texttt{student}: So I am the leader of my class. It is like the president or mayor.\\
        \textit{TSCC} & \texttt{teacher}: Oh, like a head girl?\\
        & \hspace{.6em}$\hookrightarrow$ \texttt{GPT}: That's great to hear! Being the leader of your class is a big responsibility [...] \\ 
        \midrule
        Follow-up & \texttt{ seeker:} My friends, who I also work with, are upset with me because I am leaving for a \\
        \textit{ESConv} & \phantom{\texttt{ seeker:}}  new position. I am sad that they are upset.\\
        & \texttt{support:} Ah, that can be awkward. Are you happy about the new position? \\
        & \hspace{.8em}$\hookrightarrow$ \texttt{GPT}:\hspace{0.4em}I can understand why that would make you feel down... Remember, it’s okay to [...].\\
        \midrule
        Acknowledgment & \texttt{persuadee}: I [donate] as much as I can, which is admittedly not as much as I would like to.\\
        \textit{Persuasion} & \texttt{persuader}: I know what you mean! Sometimes it is hard to find the extra time or money to help \\
        
        & \phantom{\texttt{persuader:}} those that need it.\\
        & \hspace{1.6em}$\hookrightarrow$ \texttt{GPT}: That's wonderful to hear! [...] Would you be interested in supporting Save the [...]?\\
        \bottomrule
    \end{tabular}}
    \caption{\textbf{Examples of \framework{} from our datasets,} where the act is employed by the expert from each dataset (\texttt{teacher}, \texttt{supporter}, or \texttt{persuader}). Additionally, we include  \texttt{Chat GPT-3.5}, where the model does not employ the same grounding act. }
    \label{tab:examples}
\end{table*}

\paragraph{Grounding as an Expert} To benchmark conversational grounding in LLMs, we focus on dialogue acts from the perspective of an \textit{expert listener.} Users frequently come to LLMs with a direct but potentially underspecified task~\cite{tamkin2022task, zheng2023lmsys}. In these settings, the user has privileged knowledge of their task; therefore, the onus of building common ground before providing a response lies initially with a model. More concretely, the perceived expert LLM should generate text that verifies grounding before completing a task. We therefore curate \framework{} from the perspective of an expert listener.

\paragraph{Conversational Grounding Ontologies} In curating \framework{}, we draw on prior dialogue  research. Several ontologies propose discrete dialogue acts for measuring conversational grounding. For example, \citet{clark1989contributing} propose a hierarchy of methods to achieve common ground, including the use of explicit discourse acts. They discuss relationships between acts like \textit{acknowledgment} (e.g. \textit{I understand}), and evidence of grounding in conversation. \citet{traum1992conversation} outline a range of concrete grounding acts, from simply \textit{continuing} a conversation to \textit{repairing} a misunderstanding or \textit{requesting} repair. \citet{purver2004theory} further contributes a theory of repair through \textit{clarification} requests---dialogue acts used to verify if contributions should be added to the common ground.  

 We curate a small but general subset from the large pool of proposed acts, selecting acts that are relevant to interaction with LLMs, or their current applications (e.g. teaching or emotional support). For example,  Motivational Interviewing for emotional support emphasizes asking \textit{followup} questions and signaling \textit{acknowledgment} for empathy~\cite{miller2012motivational}. Similarly, a range of pedagogical theories incentivize asking careful \textit{clarification} and \textit{followup} questions~\cite{wiske1998teaching}. And effective conflict resolution requires a careful construction of common ground~\cite{deutsch1973resolution}. We therefore select the following \textbf{three}  \framework{} that are especially relevant to these current applications of LLMs but also generalize across a range of domains where grounding is critical.

\paragraph{\texttt{\hlgraytab{Clarification}}} requests occur when a speaker seeks clarification on an utterance instead of initiating repair. Clarification is used primarily to avoid a misunderstanding, and concerns information presented in prior utterances $u_{1..i}$. In other words, clarifications serve to ``clear up'' a potential future misunderstanding (e.g. \textit{did you mean...?} or \textit{are you referring to?}), avoiding repair from a listener \cite{purver2004theory, ginzburg2001resolving,purver2003answering,purver2003means,healey2011making,healey2003experimenting,madureira2023you,kuhn2022clam,stoyanchev13,rahmani2023survey}. 

\paragraph{\texttt{\hlgraytab{Acknowledgement}}} \textit{explicitly} signals understanding (e.g. "ok," "I understand," "I see," "Yes, and", etc.). Unlike clarification/repair, acknowledgment indicates that a speaker is ready for the next relevant turn in a conversation. We only consider utterances whose sole purpose is acknowledgment (i.e. they \textit{exclusively} contain ack. markers.) ~\cite{schegloff1982discourse, sacks1978simplest, schiffrin1987discourse, clark1989contributing, cho-may-2020-grounding}

\paragraph{\texttt{\hlgraytab{Followup}}} questions ask for elaboration on a prior utterance $u$. Followups implicitly signal understanding of $u$. Unlike clarifications---which are concerned wholly with misunderstandings---followups signal understanding by seeking additional information on a prior utterance $u$. Concretely, follow-ups indicate understanding by attempting to ground on related information. While clarification is about understanding the existing interaction more thoroughly, a follow-up is concerned with continuing the current interaction\footnote{One nifty way to distinguish follow-ups and clarification questions is the "O.K." test, introduced by \citet{benotti2021recipe}. In short, if prefixing a potential clarification (e.g. "Do you mean X?") with an acknowledgment (e.g. "O.K.") sounds awkward, then it's likely a clarification.} \cite{davis1982determinants, graesser1995collaborative, traum1992conversation, bunt2017dialogue}.

\section{Data}
\label{simulation_datasets}

We choose to analyze conversations in three domains where grounding is critical and where LLMs are already used for social skill training---education, emotional support, and persuasion. Our datasets are task-oriented in nature, have multiple turns, and consist of two participants. Since our \framework{} are curated around an expert listener, each dataset also has one expert participant. To identify grounding gaps, we can simulate utterances from the expert using an LLM. We briefly outline our datasets (in English). 

For \textit{emotional support}, we use \textbf{Emotional Support Conversations (ESConv)}, a corpus of one-to-one online conversations between a help-seeker and a help-provider, collected via crowdsourcing~\cite{liu2021towards}. To analyze grounding acts in \textit{education}, we use the \textbf{Teacher Student Chatroom Corpora (TSCC)}, a collection of written conversations captured during one-to-one lessons between teachers and learners of English~\cite{caines2020teacher, caines-etal-2022-teacher}. Finally, for \textit{persuasion}, we use \textbf{Persuasion for Good}, a dataset consisting of one-to-one online conversations between a persuader and a persuadee, where the persuader solicits donations from payouts on a crowd working website \cite{wang-etal-2019-persuasion}. Due to resource constraints, we sample 100 conversations from each dataset and truncate them to the median length  (TSCC = 92, ESConv = 22 Persuasion = 20). Details on our selected datasets, sampling, and truncation are in Appendix \ref{sampling_conv}. 

\section{Classifying Grounding Acts}
\label{simulation_compare}

To analyze disparities in \framework{} between human and LLM generations, we must first classify \framework{} in our datasets. In this section, we outline the construction of test/validation sets, and our prompting setup to classify \framework{}.

\subsection{Dataset Splits and Prompting}
\paragraph{Method} We turn to classifying the \framework{} in our datasets. First, we withheld and annotated a validation (10\%) and test (10\%) set of conversations from each dataset in \S\ref{simulation_datasets}. Messages were annotated for a single \textbf{primary} act---if an utterance contained multiple \framework{}, we selected the most relevant one. Two authors participated in the annotation process. After annotating, the authors highlighted disagreements, discussed, and broke ties, yielding a final Cohen Kappa agreement of $\kappa=0.72$. Following annotation, the first author prompt-engineered a multi-class classification prompt on the validation set, using both zero-shot and few-shot prompting. We used the latest \texttt{GPT} model during the time of writing (\texttt{GPT-4}) as our classification model, with temperature = 0 but default parameters otherwise.\footnote{We
use standard parameters provided in OpenAI’s API
(max\_tokens = 256).} Queries were run between July-November 2023. 

\paragraph{Results} \texttt{GPT-4} can identify \framework{} in conversations with reasonably high accuracy (avg. Macro F-1 across datasets = \hlgraytab{\textbf{0.89}}, Appdx. Table \ref{tab:acc}). In the 0-shot setting, however, we find that \texttt{GPT-4} frequently misclassifies follow-ups and clarification questions (avg. F-1 clarification = \hlgraytab{\textbf{0.40}}; follow-up = \hlgraytab{\textbf{0.70}}). Few-shot prompting substantially increases model performance (clarification F-1 = \hlgraytab{\textbf{0.85}}; follow-up F-1 = \hlgraytab{\textbf{0.91}}).  

\section{Analyzing Grounding Acts in LLMs}
\label{sec:simulation}
Given our three \framework{}, a reasonably performing \texttt{GPT-4} classifier for labeling them, and a set of target datasets \& metrics, we can now measure disparities between humans and LM simulations. In this section, we outline our controlled simulation process (\S \ref{sec:sim_proc}) and metrics to measure conversational grounding~(\S \ref{metrics}). 

\subsection{Simulation Method}
\label{sec:sim_proc}
We start with a conversation $C_{1..N}$ consisting of $N$ ordered role utterance pairs $(r_t, u_t)$. Each of our selected datasets has two unique roles, expert and listener. Since LLMs are generally leveraged for the role of an expert listener, we focus on simulating the expert. Given this setup, our simulation process generates controlled counterfactual messages $g_t$ for each $u_t$.

Concretely, to simulate an utterance at timestep $t$, we extract all messages until $t$: $C_{1...{t - 1}}$. Then, we input this context into a selected LM (\texttt{LM($C_{1...{t - 1}}$)}), along with a high-level instruction (e.g. \textit{Roleplay a therapist}). The LM then generates the next message, offering a counterfactual to the human ground truth. Using this process, each ground truth utterance $u_t$ from our selected datasets has a controlled, LM-generated counterpart $g_t$, conditioned on the \textit{same} conversational history. Figure \ref{fig:example_interact} summarizes this process.

After generating LM counterfactuals $g_t$, we can similarly compute the rate of \framework{} across our generated messages. While we use \texttt{GPT-4} to classify \framework{}, simulating conversation for all our datasets is prohibitively expensive. In this section, we use \texttt{GPT-3.5} with default parameters for conversation simulation (later, in \S\ref{sec:ablate-instr}, we experiment with a wider range of models).

\subsection{Metrics to Measure Grounding Acts}
\label{metrics}
How should we measure the use of our selected grounding acts? We propose two metrics: base rate and Cohen's $\kappa$. The base rate provides a measure of grounding frequency; specifically, how often a human utterance or LM generation contains a specific grounding act. In contrast, Cohen's $\kappa$ measures agreement in \framework{} between humans and LMs. Just because an LM generation contains an act does not mean it occurs in the same place as in human dialogue. Our metrics apply to any dialog dataset $D$ consisting of conversations with utterances-role pairs $\{(r_1, u_1),...,(r_n, u_n)\}$. 

\paragraph{Base Rate} We first compute the overall frequency of grounding acts subset $G$ as $P(u_i \in G)$. The base rate provides a reference point for the overall presence of specific conversational grounding acts.

\begin{table}[]
    \centering
    \resizebox{\linewidth}{!}{%
    \begin{tabular}{l|rr|r}
        \toprule
        Act & \texttt{ChatGPT 3.5} & Human & Cohen $\kappa$   \\ 

        \midrule
        & \multicolumn{3}{c}{Emotional Support Conv} \\
        \midrule
        
        Follow& 10.78 $\pm$ 2.1 & 27.87 $\pm$ 4.4 & \hlsecondarytab{12.47} $\pm$ 6.4 \ \ \\
        Ack. & 1.05 $\pm$ 0.8 & 12.9 $\pm$ 3.7 & \hlsecondarytab{3.14} $\pm$ 4.9 \ \ \\
        Clar. & 0.0 $\pm$ 0.0 & 3.05 $\pm$ 1.2 & \hlsecondarytab{0.0} $\pm$ 0.0 \ \ \\

        \midrule
        & \multicolumn{3}{c}{Teacher Student Chatroom} \\
        \midrule 
Follow & 11.56 $\pm$ 1.9 & 12.04 $\pm$ 2.1 & \hlsecondarytab{16.75} $\pm$ 4.6 \  \\
Ack. & 5.68 $\pm  $ 1.4 & 16.59 $\pm$ 2.4  & \hlsecondarytab{18.25} $\pm$ 5.4 \ \\
Clar. & 0.57 $\pm$ 0.3 & 3.77 $\pm$ 0.9  & \hlsecondarytab{0.36} $\pm$ 2.5 \ \\

        \midrule 
        & \multicolumn{3}{c}{Persuasion for Good} \\
        \midrule
Follow & 1.66 $\pm$ 0.9 & 8.18 $\pm$ 2.4 & \hlsecondarytab{2.94} $\pm$ 7.6 \ \ \\
Ack. & 1.8 $\pm$ 1.0 & 6.11 $\pm$ 1.9 & \hlorangetab{25.73} $\pm$ 16.7 \\
Clar. & 0.0 $\pm$ 0.0 & 0.28 $\pm$ 0.4 & \hlsecondarytab{0.0} $\pm$ 0.0 \ \ \\
        \bottomrule
    \end{tabular}
    
    }
    \caption{\textbf{Grounding acts and associated metrics across our datasets}. $\pm$ represents 95\% confidence intervals (bootstrapped). Humans use \framework{} more than LLMs; furthermore, LLMs show \hlsecondarytab{poor} agreement (Cohen $\kappa$) with humans. 
    }
    \label{tab:overall_results}
\end{table}

\paragraph{Cohen Kappa $\kappa$} While the base rate provides a measure of frequency, it does \textit{not} capture the discrepancy between grounding acts used between a candidate and reference conversation. For instance, an LM-generated conversation $M$ may use \textit{more} grounding strategies, but only in locations that rarely match human $H$ use. Consider a simulated dialogue agent that always generates a grounding act. 
While the language this agent generates doesn't have the problem of presuming common ground, an agent that always generates grounding acts can be understandably irritating~\cite{horvitz1999principles}. 

To this end, we use \textbf{Cohen $\kappa$}~\cite{cohen1960coefficient}, a measure of inter-rater agreement bounded between -1 and 1. Cohen $\kappa$ has several useful properties: it is well-validated across social scientific studies---agreement can be compared to pre-existing work to determine strength. Furthermore, $\kappa$ adjusts for random chance: values of $\kappa < 0$ indicate that agreement is worse than chance. In measuring \framework{}, we treat $M$ and $H$ as individual raters.

\begin{table*}[]
    \centering
    \resizebox{\textwidth}{!}{
    \begin{tabular}{l|rr|rr|rr}
        \toprule
        \multirow{2}{*}{\backslashbox[50mm]{\textbf{Generator}}{\textbf{\textsc{Act}}}}
            & \multicolumn{2}{c}{\hlprimarytab{$+$ Followup}} & \multicolumn{2}{c}{\hlprimarytab{$+$ Acknowledgement}} & \multicolumn{2}{c}{\hlgraytab{$+$ Clarification}} \\
            \cmidrule(lr){2-3} \cmidrule(lr){4-5} \cmidrule(lr){6-7} & Base Rate & Cohen $\kappa$ & Base Rate & Cohen $\kappa$ & Base Rate & Cohen $\kappa$ \\ 
        \midrule
        \texttt{Human} & 27.87 $\pm$ 4.4 & \hlgraytab{36.41} & 12.89 $\pm$ 3.7 & \hlgraytab{30.12} & 3.05 $\pm$ 1.2 & \hlgraytab{48.45}$^{\ref{kappa_inflate}}$ \\
        \midrule
        \texttt{3.5-instruct-turbo} & 29.35 $\pm$ 3.3 & \hlorangetab{22.16} $\pm$ 7.7 & 1.49 $\pm$ 1.1 & \hlsecondarytab{5.5} $\pm$ 6.0 & 0.12 $\pm$ 0.2 & \hlsecondarytab{-0.22} $\pm$ 0.4 \  \\
        \texttt{3.5-turbo (ChatGPT 3.5)} & 10.76 $\pm$ 2.1 & \hlsecondarytab{12.47} $\pm$ 6.4 & 1.05 $\pm$ 0.8 & \hlsecondarytab{3.14} $\pm$ 4.9 & 0.0 $\pm$ 0.0 & \hlsecondarytab{0.0} $\pm$ 0.0 \  \\

        \texttt{4} & 12.26 $\pm$ 2.6 & \hlsecondarytab{11.04} $\pm$ 7.4 & 1.17 $\pm$ 0.8 & \hlsecondarytab{11.18} $\pm$ 7.7 & 0.35 $\pm$ 0.4 & \hlsecondarytab{-0.61} $\pm$ 0.6 \ \\
        
        \midrule
        \texttt{mistral-sft} & 20.03 $\pm$ 3.1 & \hlsecondarytab{18.48} $\pm$ 7.7 & 12.08 $\pm$ 3.4 & \hlorangetab{26.92} $\pm$ 9.6 & 0.47 $\pm$ 0.5 & \hlsecondarytab{12.39} $\pm$ 16.1  \\
        \texttt{mistral-dpo} & 15.99 $\pm$ 2.7 & \hlsecondarytab{15.25} $\pm$ 8.0 & 19.11 $\pm$ 4.1 & \hlsecondarytab{19.33} $\pm$ 7.7 & 0.93 $\pm$ 0.8 & \hlsecondarytab{9.79} $\pm$ 12.7  \\
        \midrule
        \texttt{3.5-turbo + mitigation } & 42.32 $\pm$ 4.7 & \hlorangetab{26.41} $\pm$ 6.6 & 2.54 $\pm$ 1.3 & \hlsecondarytab{5.02} $\pm$ 5.9 & 1.16 $\pm$ 0.8 & \hlsecondarytab{-1.65} $\pm$ 0.9 \ \\
        \bottomrule
    \end{tabular}
    }
    \caption{\textbf{ESConv \framework{} and metrics across model variants.}  We find that more recent OpenAI (\texttt{3.5-turbo, 4}) models use \textit{significantly fewer} \framework{} than humans; and that agreement (Cohen $\kappa$) with humans is \hlsecondarytab{poor} to \hlorangetab{fair} across all evaluated models.}
    \label{tab:results_all_data}
\end{table*}

\section{Gaps in Generating Grounding Acts}
\label{gaps_in_grounding}

Having introduced the controlled simulation process, we now report metrics on \framework{}~(\S \ref{sec:lm_results}) and qualitatively analyze errors (\S \ref{sec:lm_error_anal}).

\subsection{Simulation Results}
\label{sec:lm_results}
First, we find significant discrepancies between humans and LLMs when using \texttt{GPT-3.5} for conversation simulation (Table \ref{tab:overall_results}). Across all datasets, LLMs generations contain fewer grounding acts, like followups (avg. \hlsecondarytab{\textbf{64.3\%}} decrease) and ack. (\hlsecondarytab{\textbf{83.4\%}}) acts. Clarifications never occur when using \texttt{ChatGPT-3.5} for ESConv and Persuasion. While \texttt{ChatGPT-3.5} does initiate some clarification on TSCC, we observe a \hlsecondarytab{\textbf{84.8\%}} decrease. 

Beyond rate discrepancies, we observe low agreement between \texttt{ChatGPT-3.5} and humans---LLM generations rarely contain \framework{} in same position as human utterances. Of the 3 \framework{} $\times$  3 dataset pairs, only 3 / 9 have a Cohen $\kappa$ agreement significantly greater than \textbf{zero}, with $\kappa$ averaging \hlgraytab{\textbf{10.73}} for followup, \hlgraytab{\textbf{11.13}} for acknowledgment, \hlgraytab{\textbf{0.23}} for clarification. To confirm that human-LM agreement is low, we run a human-human study with ESConv (Appendix \ref{sec:human_kappa}), and observe significantly higher $\kappa$ across grounding cts (avg. $\kappa \approx$ 37.5).

\subsection{Error Analysis}
\label{sec:lm_error_anal}

We performed qualitative analyses to develop an in-depth understanding of how LM generations fail to incorporate grounding acts. First, we performed inductive coding to produce a set of frequent errors. An annotator (one of the authors) read a random sample of 160 instances where a human uses a grounding act, while the simulated supporter does not. Relevant turns were examined together with the previous turn for context (details in Appdx. \ref{error_analysis_appdx}).

Simulated supporters fail to use \textbf{(1) acknowledgment} to show empathy (\hlgraytab{\textbf{43.94\%}}) or ask \textbf{(2) followup} questions for more information (\hlgraytab{\textbf{26.52\%}}) / to continue a conversation (\hlgraytab{\textbf{12.88\%}}). Furthermore, simulated supporters do not ask \textbf{(3) clarification} questions to verify understanding (\hlgraytab{\textbf{9.09\%}}), or resolve a specific ambiguity (\hlgraytab{\textbf{7.58\%}}). Table \ref{tab:qual} in the Appendix contains more detail on each error type and qualitative examples. Next, we investigate potential causes of the \textit{grounding gap} and use discovered error types to design an informed prompting mitigation.

\section{Why do Grounding Gaps Emerge?}\label{sec:ablate-instr}

Here, we explore potential mechanisms for how grounding gaps emerge. We focus on analyzing ESConv, a target domain where disagreement in \framework{} is especially consequential.

First, we examine \framework{} across several OpenAI \texttt{GPT} variants and observe a larger grounding gap in \textit{newer} models~(\S\ref{sec:openai_var}). To understand the roots of this trend, we evaluate open-source models. We hypothesize that current supervised-finetuning (SFT) and preference optimization (PO) datasets drive human-LM disagreement. To test this, we rigorously isolate the effects of SFT and PO training on grounding agreement~(\S\ref{sec:sft}).

\subsection{LLM Variants}
\label{sec:openai_var}
\paragraph{Method} To further investigate the grounding gap, we test a wider range of \texttt{GPT} models, rerunning our simulation process for ESConv on \texttt{gpt-3.5-instruct} (a replacement for legacy OpenAI models, trained similarly to the \texttt{text-davinci-00X} series) and \texttt{gpt-4}. We examine \framework{} across these models.

\paragraph{Results} Generations from the OpenAI model variants have lower \framework{} base rates and poor agreement with humans ($\kappa$ < 0.2). A surprising exception to this is \texttt{3.5-instruct-turbo} for followups, where \texttt{instruct} shows \hlorangetab{fair} agreement with humans \textit{and} uses \framework{} at a similar rate ($29.5$ \texttt{instruct} vs. $27.9$ Human). While \texttt{instruct} is trained using an ``older'' procedure, it produces a better $\kappa$ / base-rate tradeoff compared to most models. 

\begin{figure}[t]
\centering
\begin{minipage}{.48\columnwidth}
    \centering
    \includegraphics[width=\linewidth]{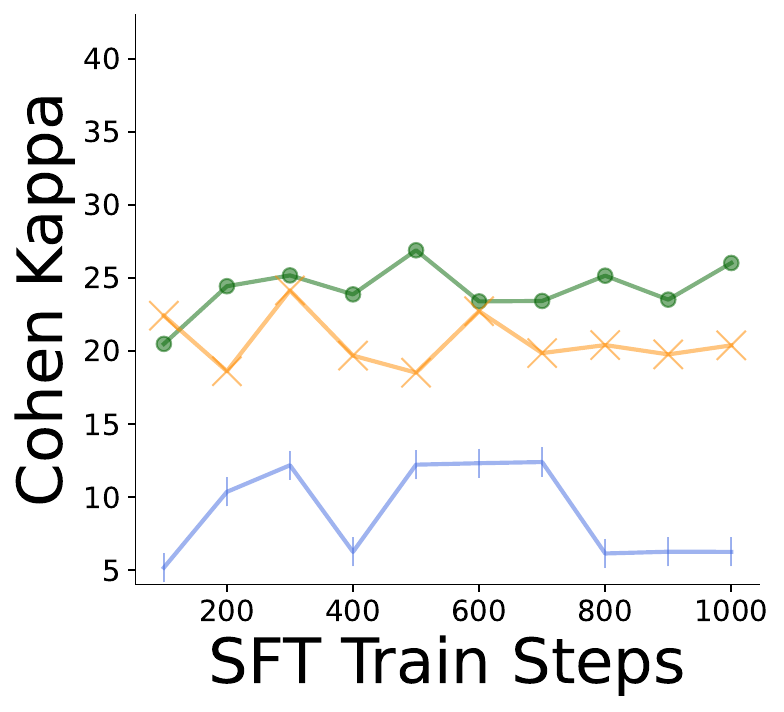}
    \subcaption{}
    \label{fig:enter-label}
\end{minipage}%
\begin{minipage}{.48\columnwidth}
  \centering
  \includegraphics[width=\linewidth]{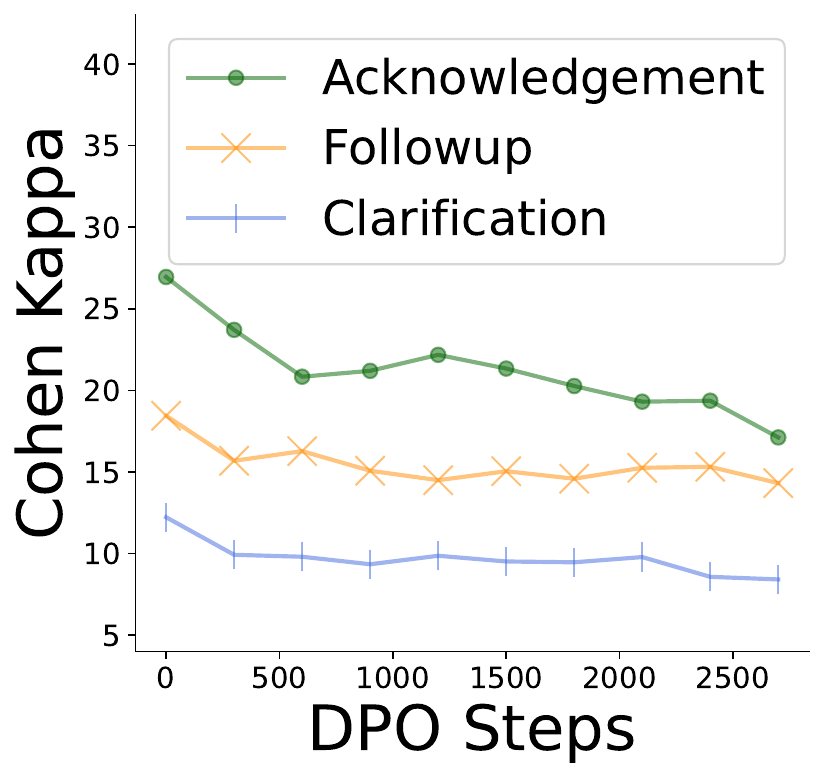}
\subcaption{}
  \label{fig:test2}
\end{minipage}
\caption{\textbf{The role of SFT and preference optimization in grounding agreement.} (a) We observe \textbf{no correlation} between \textbf{SFT} training steps and Cohen $\kappa$ agreement on \framework{}, with a Pearson R correlation test yielding insignificant results: $p > 0.1$ (b) We observe \textbf{negative correlation} between \textbf{DPO} train steps and Cohen $\kappa$ agreement on \framework{}, with Pearson R averaging $R = -0.79$, and $p < 0.05$ for all acts.}
\end{figure}

\subsection{SFT \& Preference Optimization}\label{sec:sft}

OpenAI models, however, are closed source: to isolate training procedures that impact grounding agreement, we independently evaluate the role of current SFT and PO datasets. 

\paragraph{Method} We investigate if the standard SFT + PO training setup improves use of \framework{}; and if the amount of SFT + PO matters. At a high level, we replicate the training procedure for Zephyr, an open-source instruction following LM~\cite{tunstall2023zephyr}. First, we \textbf{SFT} Mistral 7B~\cite{jiang2023mistral} for three epochs on a filtered version of the UltraChat dataset~\cite{ding2023enhancing, tunstall2023zephyr}.\footnote{For additional training details, see Appendix \ref{sec:training_deets}.} During the SFT process, we save a total of 10 evenly-spaced checkpoints across a training run. Each checkpoint is used to re-simulate conversations and measure the use of \framework{}. Next, we use preferences from the synthetic UltraFeedback~\cite{cui2023ultrafeedback} dataset for the \textbf{PO} stage. We run direct policy optimization (DPO)~\cite{rafailov2023direct} for an additional three epochs, starting with the highest $\kappa$ checkpoint on \framework{} from the prior SFT stage. Like with SFT, we save 10 evenly spaced checkpoints and rerun our simulations using each checkpoint.

Beyond synthetic datasets, we additionally evaluate the Archangel models~\cite{ethayarajh2024kto}, released to study the effect of various instruction tuning and preference optimization (both DPO and PPO~\cite{schulman2017proximal}) procedures. From the Archaengel suite, we evaluated the final Llama 7B models trained on a mixture of real human feedback datasets: OpenAssistant~\cite{kopf2024openassistant}, Stanford Human Preferences~\cite{ethayarajh2022understanding} and Anthropic HH-RLHF~\cite{bai2022training}.

\paragraph{Results} Across open-source Mistral experiments, we find no evidence that SFT impacts grounding agreement: Pearson correlation across \framework{} vs. SFT checkpoints is $\approx 0$, with $p > 0.1$. Still, while Cohen's $\kappa$ is poor throughout training $(\kappa < 0.3)$, we observe that SFT-only \texttt{mistral} has the highest $\kappa$ across acknowledgment and clarification for all evaluated models---even when including closed-source OpenAI variants.

On the other hand, increased DPO training on Mistral 7B \textbf{degrades} agreement across all \framework{}: followup ($R = - 0.70, p < 0.05$), acknowledgment ($R = - 0.89, p < 0.05$), and clarification ($R = - 0.78, p < 0.05$). Similar to how PO induces longer responses~\cite{singhal2023long}, we observe bias towards \textbf{assuming grounding} instead of employing \framework{}. Altogether, instruction following SFT \textit{does not} improve grounding agreement, and added PO erodes it. 

We observe similar degradations with PPO / DPO on the Archangel suite of models (Table~\ref{archangel-results} in Appendix). After PPO, the base-rate falls by an average of \hlsecondarytab{7.0\%} across grounding acts, and $\kappa$ decreases by \hlsecondarytab{39.0\%}. Similarly, DPO results in an average base-rate decrease of \hlsecondarytab{12.1\%}, and an average $\kappa$ decrease of \hlsecondarytab{42.5\%}. Regardless of the base model or preference optimization algorithm, we observe decreases in generating grounding acts.

We further examined our SFT and preference datasets to identify a potential source of the grounding gap. As a heuristic, we simply searched for questions in assistant responses, and found that assistant responses containing questions are overall relatively rare---11.83\% of samples in UltraChat and 18.35\% of samples in UltraFeedback have an utterance where the assistant's answer contains any question at all (followup questions and clarification questions are a subset of these). Second, we found that the UltraFeedback dataset explicitly signals that asking questions is dispreferred: questions are significantly \textbf{less frequent} in preferred (13.77\%) compared to dispreferred~(18.35\%) examples, ${\chi}^2 = 484.08$, $p<0.00001$.

\subsection{Prompting Mitigations}
\label{sec:prompt_mitigation}

Lastly, we explore a potential prompt-based intervention. We design a prompt around our qualitative error analysis and re-evaluate \framework{}. 

\paragraph{Method} 
We add a mitigation prompt (full text in Appendix \ref{appdx:mitigation}) to \texttt{Chat GPT-3.5 Turbo}, instructing it to avoid errors from our analysis (\S\ref{sec:lm_error_anal}); specifically, to \textbf{(1)} ask clarification questions when necessary, \textbf{(2)} use follow-up questions to continue a conversation, and \textbf{(3)} use acknowledgment to show empathy. Our prompting approach is similar to related work on preference elicitation~\cite{li2023eliciting}, where prompting a model to first clarify a task improves human ratings for task performance.

\paragraph{Results} Our prompt mitigation on \texttt{ChatGPT 3.5} initially looks promising, with base rates substantially higher across the board: followup nearly quadruples ($10.76 \rightarrow 42.32$), acknowledgment doubles ($1.05 \rightarrow 2.54$), and clarification increases from $0 \rightarrow 1.16$. \textbf{However}, we note that interventions result in overeager grounding: simulated supporters overuse \framework{} for minimally improved agreement (e.g. Cohen's $\kappa$). For example, while mitigation results in \textit{$50\%$ more followups than a human supporter}, agreement is still poor. Furthermore, for clarification, we find that $\kappa$ \textbf{decreases} after mitigation ($0 \rightarrow -1.65$), with mitigated $\kappa$ \textit{significantly worse than random chance.} Altogether, we find that while a prompting intervention significantly increases base rates of \framework{}, it yields minimal increases (and potentially decreases) across Cohen $\kappa$ agreement with humans, indicating that the grounding gap is a fundamental problem, difficult to address by prompting alone.

\section{Related Work}

\paragraph{Background} In operationalizing the concept of common ground \cite{clark1996using,clark1989contributing}, we build on prior work in linguistics, cognitive psychology, and communication. This includes literature on conversational structure \cite{jefferson1972side} and on subdialogues~\cite{LitAllCs87,Lit85}.

\paragraph{Conversational grounding in NLP} To benchmark grounding abilities, existing works have operationalized tasks relevant to conversational grounding, including question answering \cite{testoni2020they}, producing human-like acknowledgments \cite{paranjape2021human}, addressing ambiguities \cite{paek1999uncertainty}, providing conversational feedback \cite{pilan2023conversational,eshghi2015feedback}, addressing repair~\cite{balaraman2023thats}, asking follow-up~\cite{li2023eliciting} and clarification questions \cite{purver2004theory}. Correctly leveraging grounding strategies is particularly consequential in tasks that require coordination in order to achieve a goal \cite{bara2021mindcraft,mohanty2023transforming,fried2022pragmatics,li-boyer-2015-semantic}, play games~\cite{madureira2023instruction, shaikh-etal-2023-modeling}, plan ahead~\cite{Chu98,Loc98}, retrieve data~\cite{lu-etal-2023-statcan}, or improvise~\cite{cho-may-2020-grounding}. In such tasks, the use of grounding acts has been shown to increase success and conversation quality \cite{zhou2022reflect}. Furthermore, the ability to establish a common ground is a key component in efforts to design believable conversational agents \cite{park2022social,park2023generative,aher2022using,argyle2022out} and facilitate human-AI collaboration in dialogue~\cite{lin2023decisionoriented} Our work synthesizes existing literature by formalizing a framework to study grounding in human-AI dialogue.

\paragraph{LLMs and conversational grounding} Despite the ubiquity and importance of conversational grounding, previous work has identified fundamental limitations in the ways dialogue agents powered by large language models establish common ground, noting that current systems usually guess what the user intended, instead of leveraging grounding acts.\footnote{\url{https://openai.com/blog/chatgpt}} Related to this limitation, various undesirable conversational patterns have been identified, including over-informative question answering \cite{tsvilodub2023overinformative}, refusal to answer ambiguous questions \cite{abercrombie2023mirages,min2020ambigqa,gao2021answering}, miscalibration issues \cite{nori2023capabilities,zhou2023navigating}, and overconfidence \cite{mielke2022reducing}. Similarly, LLM sycophancy \cite{perez2022discovering}---mirroring the views of a user---may be related to presumptive grounding. Large language models' generations have thus been 
criticized as not being grounded in any communicative intent, any model of the world, or any model of the reader’s state of mind \cite{bender2021dangers}. In our work, we carefully examine the role of contemporary instruction following and preference optimization, analyzing their effect on conversational grounding.

\section{Discussion}
\begin{quote}    
    \texttt{(CAN YOU ELABORATE ON THAT)}
     \\\phantom{ab}--- \textbf{ELIZA Rule~\cite{eliza_citation}}
\end{quote}
\noindent Across evaluated models, we observed reduced rates of grounding acts and poor grounding agreement with humans (\S\ref{gaps_in_grounding}). We also isolated sources of reduced grounding act use (\S\ref{sec:ablate-instr}). Here, we reflect on findings and outline avenues for future work. 

\paragraph{On the risks of \textit{not} generating grounding acts.} Most of our evaluated LLMs generate \framework{} at a significantly lower rate than humans. Instead of initiating a grounding act, instruction following LLMs simply ``provide the answer.'' In low-stakes situations like informal conversation or chit-chat, assuming grounding may be acceptable. However, LLM simulations are used extensively across a range of critical tasks, like social skill training, where appropriate grounding is necessary. We find substantial disparities in these domains: specifically, teaching, persuasion, and therapy. Furthermore, we suspect that these disparities extend far beyond our evaluated domains, e.g. cross-cultural interaction, medical or legal advice, customer support, and beyond. 

\paragraph{On contemporary SFT and PO.} We find that current preference datasets explicitly signal that asking questions is dispreferred (\S \ref{sec:sft}). Non-expert humans have different preferences and opinions~\cite{casper2023open}, but, in single-step interaction, might agree on salient characteristics of answers including accuracy, relevance and clarity. This results in narrowly scoped preference datasets, where responses that are immediately relevant to a prompt are preferred. However, grounding by asking questions is integral in contexts such as tutoring and emotional support; failing to ground can place a burden on support-seekers. Contextualizing preferences across domains may provide a potential solution. In settings where grounding is critical, dialogue agents should prefer to use grounding acts. Still, while \framework{} provides an umbrella for grounding strategies, a closer examination of strategies routinely used by domain experts can inform preferred interaction.

\paragraph{On alignment beyond single-step interaction.} Current RLHF-trained language models are trained to optimize single-step interaction. Humans, however, strategically use \framework{} across multiple turns. While older NLP systems like ELIZA~\cite{eliza_citation}---a psychotherapy chatbot---do not explicitly model human grounding patterns, these systems still incorporate a large number of clarification and follow-up transforms. Similarly, we can augment SFT and preference datasets with \framework{}, or train reward models across multi-step interactions~\cite{hong2023zero}. However, simply using \framework{} to augment training does not guarantee that LLMs are well-aligned with humans. For example, while prompting LLMs to use \framework{} increases the use of underlying dialogue acts, it does not improve $\kappa$ agreement with humans (\S\ref{sec:prompt_mitigation}). When designing training curricula, \framework{} offer a promising direction toward measurable and grounded human-AI interaction.

\section{Conclusion}
Instruction-following language models are trained to "follow" instructions. Thus, datasets and algorithms for finetuning LLMs are designed around single-step interaction. In this work, we outline a set of discourse acts---grounding acts---to measure interaction with language models \textit{beyond} instruction following. We apply theory from conversational grounding to interactions with LLMs, finding significant differences between how humans ground in dialogue and how LLMs generate grounding acts. Designing new datasets, models, and methods, motivated by prior work on conversational grounding, will likely be necessary to minimize the grounding gap.

\section*{Limitations}
Our characterization of simulated supporters contains some anthropomorphic metaphors, which are known to be harmful as anthropomorphism in discussing technology has long been connected to dehumanization~\cite{bender2022resisting, abercrombie2023mirages, cheng-etal-2024-anthroscore}. For brevity, we discuss simulated supporters leveraging grounding acts as a way to refer to whether LLM generations contain grounding acts. 

The set of grounding acts we consider, while simple, is also not a comprehensive collection of all grounding dialogue acts used in conversation. For example, our \framework{} focus on strategies where an expert listener uses \textit{positive} grounding---negative grounding acts like model-initiated repair are out of scope. Still, we should expect that well-aligned use of \framework{} from a speaker will be correlated with a decrease in negative grounding. For example, if a speaker clarifies appropriately, then a listener should repair less. Finally, a range of grounding acts are likely subsets of our synthesized selection: paraphrasing and restating is a subset of clarification, repeating a prior utterance is an acknowledgment, etc. A finer-grained breakdown on our selected grounding acts may yield further insights---though we observe limited use even for our more general categorization. Finally, all our datasets are in English, and our collection of grounding acts is English-centered.

There also exist interaction effects between \framework{} in conversation. Our current metrics do not explicitly analyze the interaction between individual acts (e.g. does listener's acknowledgment follow a simulated speaker's clarification?). Successful goal-oriented dialogue requires an understanding of the interaction effects between \framework{}---we leave this analysis to future work.

Furthermore, our \texttt{GPT-4} based \framework{} classifier may not generalize to domains beyond our selected datasets. Analyses on other datasets may require building a new classifier or reprompting an LLM like \texttt{GPT.} Our selected datasets are also synthetic in nature, consisting of interactions between crowd workers. Validating \framework{} agreement between humans and LLMs on in-the-wild interaction is an avenue for future work.

\section*{Ethics Statement}
Measurable grounding acts introduce new directions toward improved simulations of human conversational behavior, which can be useful for training~\cite{shaikh2023rehearsal, wang2023chatgpt}. However, we strongly caution against using grounding acts to help replace a human support-provider---especially in high-risk scenarios such as therapy and education. 

Furthermore, enabling LLMs to ask numerous clarification and followup questions can be privacy-intrusive, leading support-seekers to disclose privacy-sensitive information. Using grounding acts while collecting relevant information only is an open challenge and an avenue for future work. 

Finally, we note that efficiently grounding can be harmful when the underlying goals are harmful. While we explore Persuasion for Social Good as a dataset, one can imagine settings where grounding acts are applied to persuade for more nefarious topics (e.g. political microtargeting).

\section*{Acknowledgements}
Omar Shaikh is supported by the Brown Institute's Magic Grant. Kristina Gligorić is supported by Swiss National Science Foundation (Grant P500PT-211127). This work is also funded by the Hoffman–Yee Research Grants Program and the Stanford Institute for Human-Centered Artificial Intelligence. We thank members of the Jurafsky Lab, SALT Lab, Will Held, and Eric Horvitz, for their valuable feedback on this manuscript.

\bibliography{anthology,custom}
\bibliographystyle{acl_natbib}

\appendix
\setcounter{figure}{0}    
\setcounter{table}{0}    

\section{Dataset Details}

We discuss additional details regarding our datasets. All our selected datasets are in English, and are under an open-source license (MIT, Apache, etc.) Datasets were already all appropriately anonymized.

\paragraph{Teacher Student Chatroom Corpora (TSCC)} is a collection of written conversations captured during one-to-one lessons between teachers and learners of English~\cite{caines2020teacher, caines-etal-2022-teacher}. The lessons took place in a synchronous chatroom. The dataset contains a total of 260 conversations, spread across two dataset releases. The one-to-one chatroom lessons allowed interactive, immediate, and personalized conversations. We selected the tutoring domain since numerous opportunities of current LLMs for education have been proposed in the literature, for instance, to create educational content, facilitate instruction, student engagement and interaction, provide feedback, and personalize learning experiences \cite{kasneci2023chatgpt,demszky2021can,wang2023chatgpt}. In line with prior applications of LLMs, we use \textbf{the teacher} as the expert from this corpus.

\paragraph{Persuasion for Good} consists of one-to-one online conversations between a persuader and a persuadee \cite{wang-etal-2019-persuasion}. In the data collection, one participant was asked to persuade the other to donate to a specific charity by presenting personally relevant and appealing arguments. The dataset contains 1017 conversations and was collected on a crowdsourcing platform. Participants were encouraged to continue the conversation until an agreement was reached on donating and, if so, how much. We selected the persuasion domain since the use of current LLMs for persuasion has been proposed, for instance, to facilitate conversations about politically divisive topics \cite{argyle2023ai,tappin2022quantifying}, or to generate persuasive pro-vaccination messages \cite{karinshak2023working}. In this corpus, \textbf{the persuader} is the expert.

\paragraph{Emotional Support Conversations (ESConv)} is a corpus of one-to-one online conversations between a help-seeker and a help-provider, collected via crowdsourcing~\cite{liu2021towards}. The dataset contains 1053 conversations. Before participation, help providers were trained to provide effective support through an emotional support tutorial covering support stages and concrete strategies. We selected the emotional support domain since people turn to the web and widely accessible LLMs to seek information and receive support related to their wellbeing \cite{carlbring2023new,white2009cyberchondria}. In this corpus, the \textbf{health provider} is the expert.

\section{A Human Reference Point for $\kappa$}\label{sec:human_kappa}

We observed low agreement between \texttt{ChatGPT-3.5} and the original human dialogue participants in using \framework{}.  But the original dialogue participants presumably had lots of information that the LLM might not have had. To confirm that the LLM agreement with humans is low, we would need a fairer comparison: comparing the agreement in grounding between two human crowdworkers who are both generating a next turn given the same conversational background as the language model. We set up this small parallel task with humans, using ESConv as our evaluation task. 

Similar to our LM setup, we provide our annotators with a random sample of 50 contexts $D_{1...{t - 1}}$, and ask two different annotators to independently complete the next message, using the same prompt as with the LM. Through the study description, we informed participants that their evaluations would be used to benchmark current LLMs like \texttt{ChatGPT}. Given the domain, we recruited individuals on Prolific who self-reported as (1) fluent English speakers, (2) recorded their workplace function as a Healthcare Professionals and (3) listed their employment role as a Therapist / Well-being counselor. Annotators were paid at a rate of \$12 / hour. Despite limited instruction, human-human agreement across using any grounding act is fair, with Followup $\kappa = 35.66$, Clarification $= 48.45$, and Acknowledgement $= 29.15$. All human-human $\kappa$ scores are substantially higher than Human-LM counterparts. Note that this is still effectively a \textit{lower bound} on $\kappa$ between two independently recruited individuals---with additional discussion between two mental health supporters, we would expect this score to be much higher.\footnote{\label{kappa_inflate}We note that clarification $\kappa$ might be inflated, since support is low. See Table \ref{tab:human_kappa_use} in Appendix for details.}

\section{Sampling Conversations}
\label{sampling_conv}
Because we aim to study grounding in language models, we must control for context length within datasets. With in-context learning (ICL), models may unfairly adapt to longer conversation history, learning to employ \framework{} only on longer conversations. We preprocess our dataset to control for ICL. First, we merge successive turns between the same participant into a single message.\footnote{Our selected datasets also contain successive messages from the same participant (average 23\%).} Then, within each dataset, we sample 100 random conversations greater than or equal to the median number of messages in a dataset (TSCC = 92, ESConv = 22, Persuasion = 20). Finally, we truncate the sampled conversations to the median length. 

\section{Details on Error Taxonomy}
\label{error_analysis_appdx}
We developed a taxonomy of five error types based on the notes about emerging reasons why human used grounding acts, with each error type corresponding to a specific use of a grounding act (described in Appendix, Table \ref{tab:qual}). Two authors then independently annotated the sentences, indicating whether any of the five error categories is present. We found a substantial inter-rater agreement between the two annotators (Cohen's $\kappa=0.77$). The labels from the two annotators were then aggregated such that an error category label is assigned if both of the annotators assigned it. Overall, 82.5\% of examples from the sampled set were assigned to one of the categories.

\section{Prompt Mitigation}
\label{appdx:mitigation}
We use the following prompt mitigation, designed from our qualitative analysis (\S\ref{sec:lm_error_anal}). We include this in OpenAI's \texttt{system prompt} field.
\begin{quote}
    \textit{Make sure you ask clarification questions to verify that you understand what a person is saying or to resolve an ambiguity. Also, use follow-up questions to inquire about related topics, or to continue a conversation naturally. Finally, acknowledge what a person is saying to show empathy (e.g. “o.k.“, “I understand,” etc.) when necessary. Make sure you use these strategies carefully and like a trained therapist; do not overuse them.}
\end{quote}

\section{Training Details}
\label{sec:training_deets}
During the SFT stage, we use an effective batch size of 128 (gradient accumulation) with a learning rate of 3e-4. In the preference optimization stage, we train with a learning rate of 5e-6, with a batch size of 32. To reduce memory usage, we use LoRA~\cite{hu2021lora} and train on 1 A100 GPUs. Modeling took a total of 1 day. For all experiments, we modified the HuggingFace transformers package~\cite{wolf-etal-2020-transformers}.

\begin{table*}[t]
    \centering
    \begin{tabular}{lrr}
        \toprule
        Grounding Act & Cohen $\kappa$ & Support \\
        \midrule
        Followup & 35.66 & 32 \\
        Ack. & 29.15 & 11 \\
        Clarification & 48.45 & 4 \\
        \bottomrule
    \end{tabular}
    \caption{Human-Human $\kappa$ and support for \framework{} across 100 annotations}
    \label{tab:human_kappa_use}
\end{table*}

\begin{table*}[]
    \centering
    \begin{tabular}{ll|rrr}
        \toprule
        Dataset & Grounding Act & Zero-shot & Few-shot & Support  \\
        \midrule
        TSCC & Follow-up & 0.55 & \textbf{0.91} & 47 \\
        & Ack & 0.67 & \textbf{0.89} & 65 \\
        & Clarification & 0.38 & \textbf{0.81} & 25 \\
        \midrule 
        ESConv & Follow-up & 0.90 & \textbf{0.93} & 26\\
        & Ack & 0.59 & \textbf{0.90} & 14 \\
        & Clarification & \textbf{0.83} & 0.75 & 5   \\
        \midrule
        Persuasion & Follow-up & 0.64 & \textbf{0.89} & 9 \\
        & Ack & 0.14 & \textbf{0.91} & 16 \\
        & Clarification & 0.00 & \textbf{1.00} & 2 \\
        \bottomrule
    \end{tabular}
    \caption{F-1 for \framework{} classification on a withheld test set of 10 conversations from our selected dialogue datasets. In the few-shot setting, \texttt{GPT-4} has fairly high F-1 across \framework{}.}
    \label{tab:acc}
\end{table*}

\begin{table*}[ht]
\centering
\begin{tabular}{llcc}
\toprule
Grounding Act & Source & Base Rate (\%) & Cohen Kappa ($\kappa$) \\
\midrule
\multirow{5}{*}{Followup} & Human & 27.87 & 36.41 \\
 & Llama 7B & 15.59 & 18.23 \\
 & + SFT & 15.21 & 19.12 \\
 & + SFT + DPO & 11.21 & 12.13 \\
 & + SFT + PPO & 12.12 & 13.24 \\
\midrule
\multirow{5}{*}{Acknowledgement} & Human & 12.89 & 30.12 \\
 & Llama 7B & 15.28 & 22.77 \\
 & + SFT & 16.73 & 28.56 \\
 & + SFT + DPO & 16.11 & 18.23 \\
 & + SFT + PPO & 14.05 & 19.54 \\
\midrule
\multirow{5}{*}{Clarification} & Human & 3.05 & 48.45 \\
 & Llama 7B & 3.01 & 14.47 \\
 & + SFT & 2.24 & 14.44 \\
 & + SFT + DPO & 2.11 & 5.19 \\
 & + SFT + PPO & 2.58 & 6.52 \\
\bottomrule
\end{tabular}
\caption{Grounding act metrics across the Archangel~\cite{ethayarajh2024kto} suite of models. We find that PPO / DPO results in decreases for both base-rate and $\kappa$ compared to the SFT-only model.}
\label{archangel-results}
\end{table*}

\begin{table*}[t]
\small
\begin{tabular}{p{0.25\textwidth} p{0.74\textwidth}}
\toprule
\textbf{Error type and frequency} & \textbf{Description}  \\
\midrule
\hlprimarytab{$+$ Acknowledgement} to show empathy (43.94\%)    &    Human acknowledges to show empathy, or display understanding, as opposed to going straight to suggestions and solutions.        \\
\textbf{Example:} & \texttt{seeker:} honestly im not sure. there is always that risk when doing that with guys. i better think long and hard before doing that! \\
& \texttt{supporter:} It sounds like you put a lot of thought into things which can be helpful in decision making. \\
& \hspace{.8em}$\hookrightarrow$ \texttt{GPT}:\hspace{0.4em} It's important to consider the potential consequences before taking any action. It may be helpful to evaluate what you truly want and need in a relationship before making any decisions.\\

\midrule
\hlprimarytab{$+$ Followup} to continue the conversation (12.88\%) &  Human asks a related question to keep the conversation going, typically not aiming to finish the exchange.     \\
\textbf{Example:} & \texttt{seeker:} My friends, who I also work with, are upset with me because I am leaving for a new position. I am sad that they are upset.  \\
& \texttt{supporter:} Ah, that can be awkward. Are you happy about the new position?  \\
& \hspace{.8em}$\hookrightarrow$ \texttt{GPT}:\hspace{0.4em} I can understand [...]. Communication is key in maintaining healthy relationships.\\
\midrule
\hlprimarytab{$+$ Followup} to inquire further (26.52\%) &  Human asks a closely related followup question to collect more information relevant to the topic, typically not aiming to finish the exchange.        \\
\textbf{Example:}  & \texttt{seeker:} I don't have close friends because I don't trust people  \\
& \texttt{supporter:} You said you generally don't trust family or people in general? \\
& \hspace{.8em}$\hookrightarrow$ \texttt{GPT}:\hspace{0.4em} I understand that feeling. [...] Take your time, and remember that trust is a two-way street. It requires both vulnerability and the willingness to give others a chance to earn your trust.\\
\midrule
\hlprimarytab{$+$ Clarification} to verify understanding (9.09\%) &  Before providing specific suggestions, the human restates their understanding in the form of a clarification question, as opposed to assuming correctness and providing suggestions (presumptive grounding).     \\
\textbf{Example:} & \texttt{seeker:} i was cheated on severally, and it resulted me to end the relationship  \\
& \texttt{supporter:} It sounds like you had to deal with a partner recently cheating and a breakup, is this right?  \\
& \hspace{.8em}$\hookrightarrow$ \texttt{GPT}:\hspace{0.4em} I'm sorry to hear that you've been through such a difficult experience [...] \\
\midrule
\hlprimarytab{$+$ Clarification} to resolve an ambiguity (7.58\%)  &   Human resolves any ambiguities related to a specific entity, as opposed to assuming correctness and providing suggestions (presumptive grounding).    \\
\textbf{Example:} & \texttt{seeker:} Sometimes abbreviated as 121   \\
& \texttt{supporter:} Hmm, it's fine, to be honest. At [anonymized] they were always called one-to-one classes. Or do you mean 121 is not formal? \\
& \hspace{.8em}$\hookrightarrow$ \texttt{GPT}:\hspace{0.4em} "121" is more commonly used in informal settings or casual conversations. In official or formal contexts, it's better to use the full term "one-to-one." \\
\bottomrule
\end{tabular}
\caption{\textbf{Qualitative analyses.} Detailed descriptions of error types, their frequency, and example conversations.}
\label{tab:qual}
\end{table*}

\end{document}